\pdfoutput=1

\documentclass[11pt]{article}

\usepackage[preprint]{acl}

\usepackage{times}
\usepackage{latexsym}
\usepackage{url}
\usepackage{hyperref}
\usepackage{tabularx}
\usepackage{tcolorbox}
\usepackage{graphicx}  
\usepackage{graphicx}   
\usepackage{booktabs}   
\usepackage{multirow}
\usepackage{enumitem}
\usepackage{adjustbox}
\usepackage{tcolorbox}
\usepackage{booktabs} 
\usepackage{amsfonts}
\usepackage[T1]{fontenc}
\usepackage[utf8]{inputenc}
\usepackage{microtype}
\usepackage{inconsolata}
\usepackage{graphicx}
\usepackage{xspace}
\usepackage{amsmath}

\setlist{leftmargin=*,nosep}

\title{IBPS: Indian Bail Prediction System}

\author{Puspesh Kumar Srivastava$^{1*}$ \quad Uddeshya Raj$^{1*}$\quad Praveen Patel$^{1*}$, \\
\textbf{Shubham Kumar Nigam}$^{1* \dagger}$ \quad
\textbf{Noel Shallum}$^{2}$ \quad \textbf{Arnab Bhattacharya}$^{1}$\\
$^{1}$ IIT Kanpur, India \quad
$^{2}$ Symbiosis Law School Pune, India \\
\texttt{\{puspeshk24, uddeshya24, praveenp24, arnabb\}@iitk.ac.in} \\
\texttt{shubhamkumarnigam@gmail.com} 
\quad \texttt{noelshallum@gmail.com}
}

\date{24-07-2025}

\begin{document}
\maketitle

{
\renewcommand{\thefootnote}{$*$}
\footnotetext{These authors contributed equally to this work}
\renewcommand{\thefootnote}{$\dagger$}
\footnotetext{Corresponding author}
\renewcommand{\thefootnote}{\arabic{footnote}}
}

\begin{abstract}
Bail decisions are among the most frequently adjudicated matters in Indian courts, yet they remain plagued by subjectivity, delays, and inconsistencies. With over 75\% of India's prison population comprising undertrial prisoners, many from socioeconomically disadvantaged backgrounds, the lack of timely and fair bail adjudication exacerbates human rights concerns and contributes to systemic judicial backlog. In this paper, we present the Indian Bail Prediction System (IBPS), an AI-powered framework designed to assist in bail decision-making by predicting outcomes and generating legally sound rationales based solely on factual case attributes and statutory provisions. We curate and release a large-scale dataset of 150,430 High Court bail judgments, enriched with structured annotations such as age, health, criminal history, crime category, custody duration, statutes, and judicial reasoning. We fine-tune a large language model using parameter-efficient techniques and evaluate its performance across multiple configurations, with and without statutory context, and with RAG. Our results demonstrate that models fine-tuned with statutory knowledge significantly outperform baselines, achieving strong accuracy and explanation quality, and generalize well to a test set independently annotated by legal experts. IBPS offers a transparent, scalable, and reproducible solution to support data-driven legal assistance, reduce bail delays, and promote procedural fairness in the Indian judicial system.
\end{abstract}


\section{Introduction}

India’s criminal justice system is under immense strain, with bail-related proceedings constituting a significant share of the case backlog in lower courts. As of December 2022\footnote{\href{https://www.pib.gov.in/PressReleaseIframePage.aspx?PRID=2003162}{NCRB Report, 2022}}, more than {75\%} of India’s prison population comprises undertrial prisoners, individuals not yet convicted of any offence. This alarming statistic is symptomatic of a larger systemic inefficiency, where bail applications, though intended as a swift relief mechanism, often face substantial delays in adjudication. In parallel, the country’s overall incarceration rate has surpassed {131\%} of the sanctioned prison capacity, reflecting a crisis that disproportionately affects individuals from socioeconomically disadvantaged communities.

The root of this issue lies in the burdened judicial pipeline: bail hearings, especially in subordinate courts, are among the most common legal proceedings. According to recent estimates, a large proportion of criminal case pendency at the district and high court levels arises from bail applications. Despite their frequency and routine nature, these hearings often experience substantial delay. For instance, the median disposal time for bail in some High Courts like Jammu \& Kashmir is {156 days}, while even the best-performing courts often take over three weeks on average for regular bail and more than a month for anticipatory bail. Such delays not only extend unjust pretrial incarceration but also intensify court pendency and procedural bottlenecks.

In response to this critical challenge, we introduce the \texttt{Indian Bail Prediction System (IBPS)}, a comprehensive AI-based framework designed to assist in the interpretation and adjudication of bail cases. IBPS is motivated by the urgent need for scalable, transparent, and legally grounded decision-support tools that can aid courts and legal professionals in expediting routine bail applications. By leveraging structured case attributes, statutory context, and explainable rationale generation, IBPS aspires to complement human judgment rather than replace it, thereby preserving judicial discretion while enhancing efficiency.

Our work makes several contributions. First, we present a large-scale curated dataset of over {150,000} Indian High Court bail judgments, annotated with key features such as facts, statutes, past records, health conditions, and outcomes. This represents the largest factual dataset in this legal subdomain. Second, we develop a suite of LLM-based models, including fine-tuned variants and Retrieval-Augmented Generation (RAG) setups, that predict the outcome of a bail decision (granted/rejected) and generate human-readable justifications grounded in legal reasoning. Third, we conduct rigorous evaluation using both automatic metrics and expert-assessed annotations, establishing strong baselines for factual and statute-aware bail judgment prediction.

Ultimately, IBPS aims to reduce judicial burden, promote procedural fairness, and advance access to justice through responsible AI intervention. By enabling timely, explainable, and legally consistent decision support for routine bail matters, this work lays the foundation for future legal-AI systems in India’s high-stakes judicial landscape.

\paragraph{Our Contributions:}

\begin{itemize}
    \item We curate and release the largest fact-based bail judgment dataset in the Indian legal system, comprising over 150,000 High Court bail cases annotated with structured attributes such as statutes, health status, past criminal records, custody duration, and judgment outcomes.
    
    \item We design and implement the IBPS, a framework that leverages instruction-tuned large language models to perform both outcome prediction and rationale generation, supported by legal context through RAG.
    
    \item We introduce a test set manually annotated by legal experts, enabling fine-grained evaluation of our model’s predictive and explanatory capabilities against human performance.
    
    \item We conduct a detailed experimental analysis across six configurations, evaluating both factual accuracy and reasoning quality using automatic metrics, and demonstrate that models fine-tuned with statutory context significantly outperform baselines.
\end{itemize}

To ensure reproducibility and encourage further research, the dataset and model code will be made publicly available soon.

\section{Related Work}
\label{sec:related_work}

The intersection of Artificial Intelligence and law (AI4Law) has rapidly evolved over the past two decades, fueled by the digitization of legal records and advancements in natural language processing (NLP). Legal AI research spans multiple tasks such as legal judgment prediction (LJP), statute retrieval, legal question answering, contract analysis, and legal summarization. Among these, LJP has received significant attention for its ability to support judicial decision-making. Early efforts in this domain began with statistical and rule-based models, notably the work of \citet{aletras2016predicting} on the European Court of Human Rights (ECHR). Subsequent benchmarks like EURLEX57K \citep{chalkidis2019large}, CAIL2018 \citep{xiao2018cail2018}, SwissJudgment \citep{niklaus2021swiss}, JTD \citep{yamada2024japanese}, and LegalBench \citep{guha2023legalbench} have enabled large-scale legal reasoning tasks across jurisdictions. However, most such datasets are built for civil law systems, limiting their adaptability to common law jurisdictions like India. Indian legal texts present unique challenges, length, linguistic complexity, and a reliance on precedent, yet remain under-resourced in Legal NLP. While studies like \citet{zadgaonkar2021overview}, \citet{10065229}, and \citet{sharma2021predicting} applied classical ML models to Indian legal cases, they often lacked expert annotations and interpretability. In the U.S., a substantial body of research has examined bail and pretrial detention outcomes \citep{sacks2015sentenced, demuth2004impact}, highlighting their long-term consequences on conviction and sentencing. Yet, this critical decision point remains underexplored in India. Our work addresses these gaps by introducing an expert-annotated dataset for Indian bail judgments and combining judgment prediction with rationale generation. Furthermore, recent advances in large language models (LLMs), such as BERT, LLaMA, and GPT, have demonstrated impressive capabilities in legal text understanding and generation \citep{chalkidis2021lexglue, ye2018interpretable}, though concerns around hallucination, factual grounding, and legal coherence persist. 

\section{Task Description}

Our research focuses on the task of {Bail Prediction}, which comprises two sequential components: (i) predicting the outcome of a bail application, and (ii) generating an explanation to justify the predicted outcome. This task reflects a realistic legal scenario where AI systems must not only make decisions but also provide human-understandable reasoning, thereby improving transparency and aiding legal practitioners. The task spans multiple types of bail-related applications under the Indian legal system, including \textit{Regular Bail}, \textit{Anticipatory Bail}, and \textit{Bail Cancellation} appeals.

\noindent
\textbf{Prediction Task:} The first component involves determining whether the bail is granted or not, given the details of the case. Each bail case contains a combination of factual narratives, legal charges, prior criminal records, medical and personal conditions of the accused, and legal arguments presented by the defense and prosecution. While many cases involve multiple accused, this task considers prediction for a single accused at a time.

\noindent
\textbf{Formal Definition:} Let $\mathcal{D}_R$, $\mathcal{D}_A$, and $\mathcal{D}_C$ be the sets of documents corresponding to \textit{Regular Bail}, \textit{Anticipatory Bail}, and \textit{Bail Cancellation} cases, respectively. Given a document $D \in \mathcal{D}_R \cup \mathcal{D}_A \cup \mathcal{D}_C$, the goal is to predict a binary outcome $y \in \{0, 1\}$, where:
\begin{itemize}
    \item $y = 0$: denotes \textit{``Bail not granted''} if $D \in \mathcal{D}_R \cup \mathcal{D}_A$, and \textit{``Bail not cancelled''} if $D \in \mathcal{D}_C$.
    \item $y = 1$: denotes \textit{``Bail granted''} if $D \in \mathcal{D}_R \cup \mathcal{D}_A$, and \textit{``Bail cancelled''} if $D \in \mathcal{D}_C$.
\end{itemize}

\noindent
\textbf{Explanation Task:} The second component is to generate a natural language explanation for the predicted outcome. This explanation should cite relevant factual and legal grounds present in the document and reflect the reasoning typically found in judicial decisions. This component is particularly important for improving trust and interpretability in automated legal decision-making systems.

\section{Data Preparation}
\label{sec:data_preparation}

\subsection{Raw Data Collection}

We began with the Daksh database\footnote{\href{https://database.dakshindia.org}{Daksh database}}, which provides metadata on bail cases from 15 Indian High Courts, including CNR numbers, case numbers, statutes, filing and judgment dates, and bail outcomes. This metadata covered 927,897 cases, and we used it to extract corresponding full-text judgments from the eCourts High Court portal\footnote{\href{https://hcservices.ecourts.gov.in/hcservices/main.php}{eCourts High Court portal}} using custom Python scripts.

From these, we selected five High Courts, Bombay, Kerala, Allahabad, Chhattisgarh, and Jharkhand, based on volume and balance across bail categories. For each selected case, we retrieved the most recent and complete order if multiple were listed. In total, we downloaded 208,983 bail judgments covering a diverse set of offenses, from petty crimes to serious offenses like murder, rape, and cybercrimes. Additionally, we scraped official PDF documents from India Code\footnote{\href{https://www.indiacode.nic.in}{India Code}} for Indian Penal Code (IPC), Criminal Procedure Code (CrPC), and Central/State Acts to support legal context retrieval in our RAG-based systems.


\subsection{Feature Extraction}

Judgment documents are unstructured and vary significantly across jurisdictions, time periods, and even individual judges. From each document, we aimed to extract the following features: \textit{statutes}, \textit{factual narrative}, \textit{legal arguments}, \textit{past criminal record}, \textit{health condition}, \textit{case outcome}, \textit{reasoning}, and \textit{custody duration} (calculated from arrest and judgment dates).

\paragraph{Evaluation of NLP Models.}  
We initially explored classical NLP models for named entity recognition and rhetorical role extraction:
\begin{itemize}
    \item en\_legal\_ner\_trf~\cite{kalamkar-etal-2022-named}
    \item BiLSTM-CRF and MTL models trained on LegalSeg~\cite{nigam2025legalsegunlockingstructureindian}
    \item Our SparkNLP-based custom NER model using the InLegalNER\footnote{\href{https://huggingface.co/datasets/opennyaiorg/InLegalNER}{InLegalNER}} dataset
    \item Rule-based Python scripts using fuzzy matching and regex
\end{itemize}
As detailed in Table~\ref{tab:feature_ext_models}, none of these models met our performance expectations, especially on shorter bail judgments.


\paragraph{Evaluation of LLMs.}  
We experimented with multiple open-source LLMs for one-shot and few-shot extractions over 100 manually verified cases. Inference was run on an NVIDIA L40S GPU (45GB VRAM). Table~\ref{tab: LLM_comparision_table} shows the results of this evaluation.

\begin{table}[h]
  \centering
  \small
  \setlength{\tabcolsep}{4pt}
  \resizebox{\columnwidth}{!}{
    \begin{tabular}{|l|c|c|}
      \hline
      \textbf{Model} & \textbf{Inference Time (s)} & \textbf{GPT-eval Score (/10)} \\
      \hline
      google/gemma-3-12b-it                    & 188   & 6.81 \\ 
      meta-llama/Llama-3.1-8B-Instruct         & 70    & 6.89 \\ 
      microsoft/phi-4                          & 101   & 7.56 \\ 
      mistralai/Mistral-7B-Instruct-v0.3       & 136   & 6.95 \\ 
      deepseek-ai/DeepSeek-R1-0528-Qwen3-8B    & 139   & 6.09 \\ 
      bharatgenai/Param-1-2.9B-Instruct        & ---   & ---        \\ \hline
    \end{tabular}
}
  \caption{Comparison of LLMs on feature extraction quality and efficiency}
  \label{tab: LLM_comparision_table}
\end{table}

Phi-4 emerged as the most balanced model in terms of accuracy, format consistency, and speed. It was the only model that consistently returned well-structured outputs aligned with our extraction schema. Other models, including LLaMA-3 and Mistral, performed reasonably well, but often failed to adhere to formatting or hallucinated outputs. Due to its superior performance, Phi-4 was selected as the backbone model for large-scale feature extraction.We had also tried PARAM model but it was constantly crashing due to less context size (2048 tokens) which was causing it to crash while processing longer cases.

\paragraph{Extraction Strategy.}  
We tested zero-shot, one-shot, and few-shot prompting strategies. Zero-shot prompts resulted in the poorest performance, producing incomplete or misformatted outputs. One-shot prompting, which included a structured example with the case text, produced significantly more accurate and reliable extractions. Few-shot prompting marginally improved quality but was constrained by context length (especially for long judgments) and higher token costs. Thus, we adopted the one-shot strategy for all large-scale extractions.

\subsection{Data Cleaning}

We processed metadata for over 350,000 cases, of which approximately 250,000 had judgment texts available on the eCourts portal. After eliminating documents lacking factual details (e.g., those with only a binary grant/reject order), we extracted structured information from 208,293 documents. Documents missing critical elements like incident details, statutes, judgment reasoning, or outcomes were also discarded, resulting in a final usable dataset of 150,430 records. 
%
Once the LLM-generated structured outputs were available, we parsed them using regular expressions and custom Python scripts. Final features were validated and stored in standardized JSON format for downstream tasks.

\subsection{Test Dataset Creation}
To evaluate our models against expert-verified ground truth, we constructed a high-quality test dataset through manual legal annotation. We randomly sampled 100 cases from the cleaned dataset, ensuring balanced representation across the five selected High Courts and all three categories of bail applications, Regular Bail, Anticipatory Bail, and Bail Cancellation.

Three legal experts were engaged to annotate each case. For each judgment, they independently extracted the same structured features used in model training: application type, age, health condition, criminal history, statutes, arguments, case outcome, judicial reasoning, arrest and judgment dates. The experts were also asked to clearly state the predicted outcome and provide a rationale, mimicking the expected LLM output format.
This curated subset serves as our gold-standard {\textit{Test Dataset}}, allowing us to evaluate both the prediction accuracy and the quality of generated rationales through comparison with expert-provided responses. 

\section{Dataset Analysis}
\label{sec:dataset}

This study is based on a comprehensive dataset of bail judgments collected from five major High Courts in India, Chhattisgarh, Bombay, Jharkhand, Kerala, and Allahabad, comprising a total of 208,292 bail case records. After rigorous cleaning and filtering for completeness and usability, we retained 150,430 judgments for analysis. Figure~\ref{fig:case_distribution_courts} shows the distribution of cases across these courts.The complete statistics of the dataset can be found in Table ~\ref{tab: dataset_stats}.


\begin{table}[t]
\centering
\scriptsize                         
\setlength{\tabcolsep}{3pt}         
\renewcommand{\arraystretch}{1.05}  
\resizebox{\columnwidth}{!}{        
\begin{tabular}{lccc}
\toprule
\textbf{Statistic} & \textbf{Train} & \textbf{Val.} & \textbf{Test} \\ \midrule
Total \# cases                     & 120,345 & 15,042 & 15,043 \\
Total \# sentences                 & 1,465,310 & 182,491 & 183,283 \\
Avg.\ sentences / case             & 12.18 & 12.13 & 12.18 \\
Avg.\ tokens / sentence            & 28.19 & 28.26 & 28.15 \\ \midrule
\multicolumn{4}{c}{\textbf{Sentence Count per Label}} \\ \midrule
Details of the incident            & 350,499 & 43,778 & 43,894 \\
Arguments supporting application   & 331,748 & 41,123 & 41,580 \\
Arguments opposing application     & 192,479 & 24,249 & 23,873 \\
Bail conditions                    & 343,967 & 43,000 & 43,168 \\
Reasoning                          & 246,617 & 30,341 & 30,768 \\
Statutory Context                          &  9,12,204 & 1,14,227 & 1,13,061  \\ \midrule
\multicolumn{4}{c}{\textbf{Avg.\ Tokens per Label}} \\ \midrule
Details of the incident            & 29.08 & 29.10 & 29.16 \\
Arguments supporting application   & 24.69 & 24.82 & 24.64 \\
Arguments opposing application     & 18.32 & 18.42 & 18.35 \\
Bail conditions                    & 36.72 & 36.78 & 36.51 \\
Reasoning                          & 27.43 & 27.48 & 27.30 \\ 
Statutory Context                          & 106.49 & 107.02 &  106.07 \\ \bottomrule
\end{tabular}}
\caption{Dataset Statistics}
\label{tab: dataset_stats}
\label{tab:dataset_stats}
\end{table}

The final dataset contains 13 structured variables, including case number, bail type, outcome, withdrawal status, age, health condition, criminal history, involved statutes, arrest and judgment dates, judgment length, detailed case facts, and derived crime category. We preprocessed the data by replacing missing values (e.g., \texttt{none}, \texttt{nan}) with \texttt{NaN}, converting age into numeric and categorical formats, calculating custody duration from date differences, and parsing statutes. Furthermore, we used FLAN-T5 to semantically classify crimes (e.g., theft, murder, fraud) based on case details. While some attributes were incomplete, only 34.2\% had valid age and 28.6\% had arrest dates, the dataset provided sufficient information to uncover robust patterns.

The age of the accused ranged from 11 to 95 years, with a mean of 37.3 and standard deviation of 13.1. The most represented group was aged 30–50, followed by 18–30 and 50–65. A noteworthy pattern emerged linking age to bail outcomes. As shown in Appendix Figure~\ref{fig:age_outcome_disparities}, older individuals had significantly higher bail grant rates, with applicants aged 65 and above receiving bail in 84.1\% of cases compared to only 67.7\% for the 18–30 age group.

The dataset includes three major types of bail: Anticipatory (53.7\%), Regular (45.3\%), and Bail Cancellation (1.0\%). Figure~\ref{fig:bail_type_distribution} illustrates the proportion of each category. Regular bail applications exhibited the highest grant rate at 76.7\%, while anticipatory bail stood at 72.9\%. Bail cancellations had a distinct distribution, with 60.7\% resulting in “not cancelled” decisions, as detailed in Figure~\ref{fig:bail_type_outcome}.



A particularly influential variable was criminal history. Applicants with no prior record had a 74.6\% success rate, while those with a record received bail in only 50.8\% of cases, a 23.8 percentage point drop, as shown in Figure~\ref{fig:past_record_outcome}.

\begin{figure}[ht]
  \centering
  \includegraphics[width=\columnwidth]{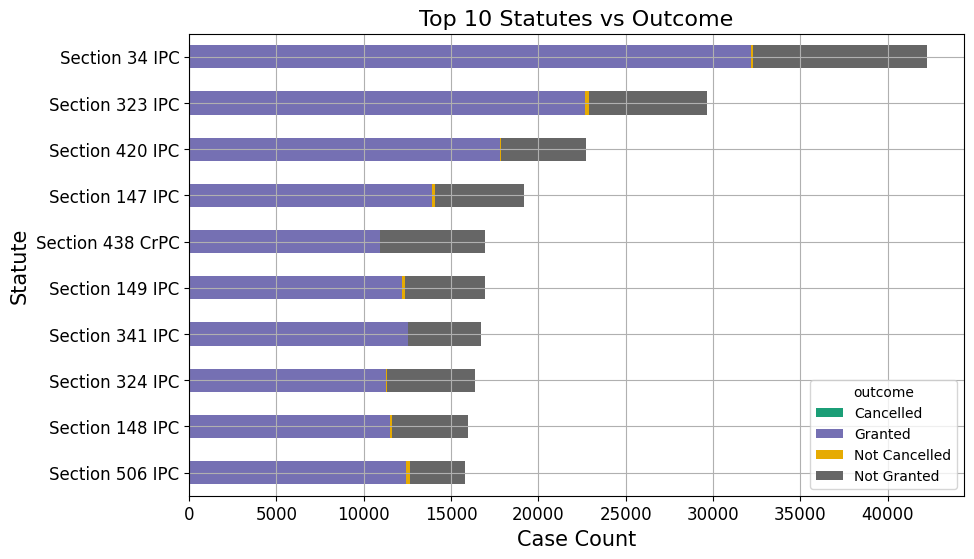}
  \caption{Top statutes and their corresponding bail outcomes.}
  \label{fig:statues_outcome}
\end{figure}

Examining statute-wise bail patterns revealed that frequently cited provisions like Section 506 (criminal intimidation), Section 420 (cheating), and Section 34 (common intention) had high grant rates (above 75\%), as depicted in Figure~\ref{fig:statues_outcome}. Conversely, socially sensitive statutes such as Section 18 of the SC/ST Act and Section 64 of the Abkari Act had some of the lowest or even zero grant rates (see Appendix Figure~\ref{fig:statutes_low_grant_appendix}).

\begin{figure}[ht]
  \centering
  \includegraphics[width=\columnwidth]{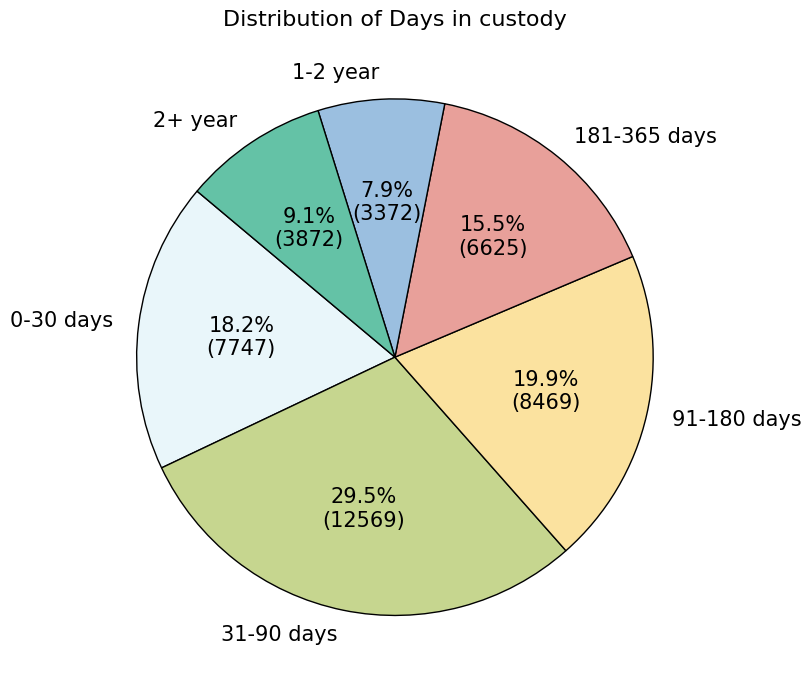}
  \caption{Custody duration distribution before judgment.}
  \label{fig:custody_duration}
\end{figure}

Temporal analysis based on 42,654 cases with complete arrest and judgment dates revealed substantial delays in pretrial detention. The average custody duration was approximately 254.6 days (median: 100 days), with a maximum of over 17 years in one extreme case (Figure~\ref{fig:custody_duration}).

\begin{figure}[htbp]
  \centering
  \includegraphics[width=\columnwidth]{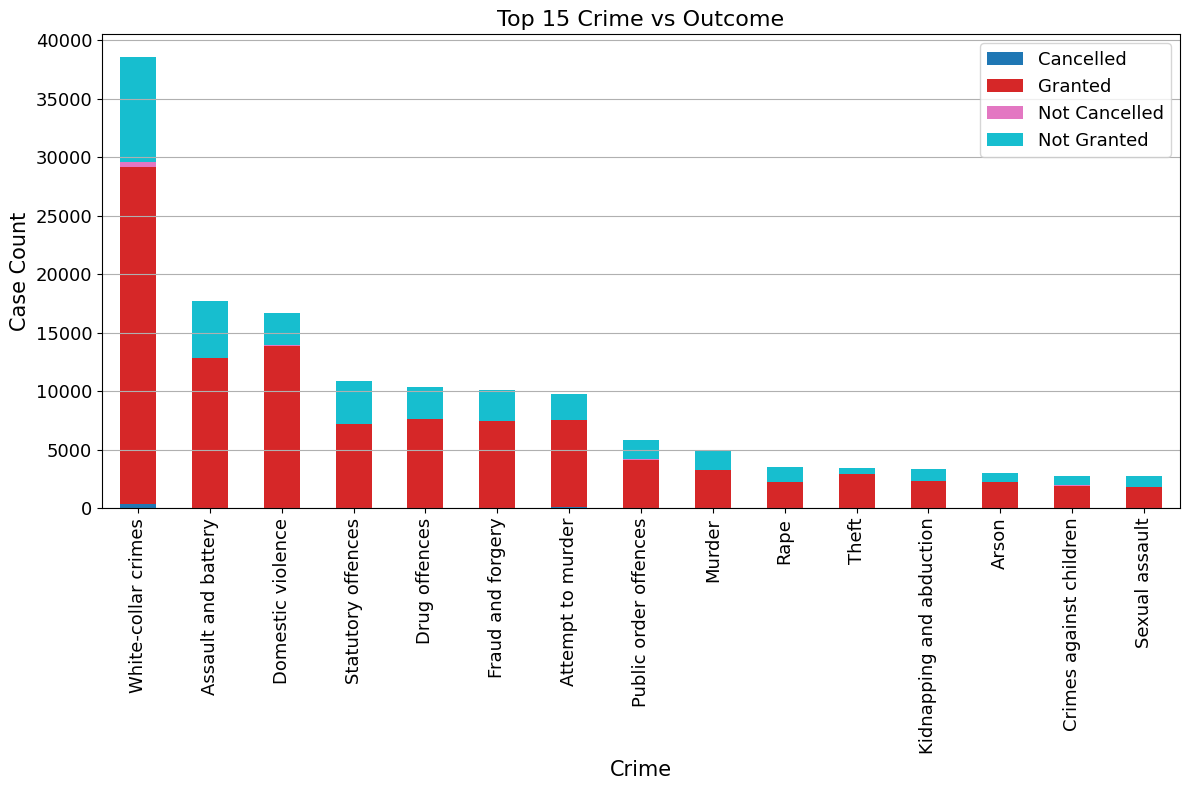}
  \caption{Bail outcomes across crime categories.}
  \label{fig:crime_outcome}
\end{figure}

Crime-wise outcome analysis (Figure~\ref{fig:crime_outcome}) showed that non-violent crimes such as theft and domestic violence had high grant rates (83.5\% and 82.9\% respectively), whereas violent crimes like rape and murder had significantly lower success rates, 62.4\% and 65.2\% respectively. Appendix Figure~\ref{fig:crime_age_appendix} further illustrates the age distribution across different crime categories, revealing that youth were disproportionately involved in crimes like kidnapping and assault, while middle-aged individuals dominated in fraud and drug-related offenses.

Finally, custody duration varied significantly by crime type. Figure~\ref{fig:crime_custody} shows that murder cases had the longest pretrial detentions, with 24.5\% of such cases extending beyond two years, while drug offenses were typically resolved within 30–90 days.


In rare instances, bail applications were withdrawn, only 1.3\% of cases (2,020 out of 150,430), suggesting that most applications proceed to judicial determination rather than being dropped voluntarily.

\section{Methodology}
\label{sec:methodology}

This section describes the design of our fine-tuning and evaluation pipeline for the task of bail judgment prediction and rationale generation. Our aim is to assess how different levels of legal knowledge and supervision influence the decision-making and explanatory capabilities of LLMs.

\subsection{Model Fine-Tuning Strategy}

We base all experiments on \texttt{microsoft/phi-4}, a compact yet highly capable LLM. Using the PEFT framework with QLoRA, we fine-tune two variants of this model:
\begin{itemize}
    \item \textbf{FT-1 (Case-Aware Model)}: This variant is fine-tuned using structured case data as input and is trained to output both the bail outcome and a natural language explanation.
    \item \textbf{FT-2 (Case+Statute-Aware Model)}: In addition to structured case data, this variant also receives textual descriptions of the applicable statutes (retrieved from India Code) during training. The model is jointly optimized to generate the outcome, explanation, and optionally reproduce the legal context.
\end{itemize}

These fine-tuned models are tested with and without RAG-based statutory context during inference. Thus, we evaluate six distinct configurations described below.

\subsection{Experimental Configurations}

\textbf{Setup 1: Baseline (Base Model, No RAG)}\\
We use the unmodified base \texttt{phi-4} model with only the structured case input, no fine-tuning, no statute context.  
\textit{Significance:} This setting tests the general legal reasoning ability of the pre-trained model in a zero-shot setting. It serves as the foundational reference point to measure gains from both fine-tuning and retrieval augmentation.

\textbf{Setup 2: Base Model + RAG (Statutory Context)}\\
The same base model is tested, but with the addition of retrieved statutory context via an RAG pipeline.  
\textit{Significance:} This setup evaluates whether access to legal definitions and statutory language improves prediction and explanation in a zero-shot setting. It helps determine the base model's ability to integrate external legal knowledge on the fly.

\textbf{Setup 3: FT-1 (Fine-Tuned on Case Data)}\\
The FT-1 model is trained using only structured case data as input. No statute definitions are used in training or inference.  
\textit{Significance:} This configuration tests the LLM's ability to generalize from past case patterns and learn case-specific priors without requiring legal text understanding. It isolates the effect of case-only training.

\textbf{Setup 4: FT-2 (Fine-Tuned with Statutory Context)}\\
The FT-2 model is trained on both structured case data and the accompanying statutory descriptions. It is evaluated without RAG at inference.  
\textit{Significance:} This setup assesses whether explicit legal knowledge, embedded during training, enhances the model's predictive accuracy and its ability to generate legally sound explanations. It also helps verify if statute-aware training improves interpretability.

\textbf{Setup 5: FT-1 + RAG (Case-Aware, Inference with Statutes)}\\
The FT-1 model (trained without statutes) is tested with retrieved statutory definitions at inference time via RAG.  
\textit{Significance:} This tests whether a case-trained model can effectively leverage unseen legal context during inference, even if it was not part of its training data. It serves as a hybrid setting between fine-tuning and retrieval-based augmentation.

\textbf{Setup 6: FT-2 + RAG (Case+Statute-Aware, RAG)}\\
The most comprehensive configuration: FT-2 (trained with statutes) is tested with statute context via RAG at inference time.  
\textit{Significance:} This setup combines learned legal knowledge and retrieval-based support to simulate the most realistic, high-performance setting. It helps determine whether retrieval complements or overlaps with learned statute representations.

\subsection{Confidence Estimation Mechanism}

To quantify the model’s certainty in its binary decision (\texttt{0} = reject, \texttt{1} = grant), we extract the generation probabilities for both possible tokens. These are normalized to sum to 100, producing an interpretable confidence score. This score is used both to compare models and to assess reliability in real-world usage scenarios.

\subsection{Prompting Schema for Structured Extraction}

Each input example during fine-tuning and inference uses a carefully designed one-shot prompt to structure the raw bail judgment text into a JSON format. This format captures key fields such as bail type, health issues, past criminal records, statutes, arguments, outcome, reasoning, and custody dates.

We place the complete prompt template in Appendix~\ref{tab: prompt} for reproducibility and reusability across models and tasks.

\section{Evaluation Metrics}
\label{sec:performance_metrics}
\begin{table*}[t]
\centering
\resizebox{\textwidth}{!}{
\begin{tabular}{|l|cccc|cccc|ccc|}
\hline
\multicolumn{1}{|c|}{\textbf{Model}} 
& \multicolumn{4}{c|}{\textbf{Outcome Metrics}} 
& \multicolumn{4}{c|}{\textbf{Reasoning Metrics}} 
& \multicolumn{3}{c|}{\textbf{Bail Conditions Metrics}} \\
& Accuracy & Precision & Recall & F1-Score 
& ROUGE-L & BLEU & METEOR & BERTScore 
& BLEU & METEOR & BERTScore \\
\hline
VANILLA 
& 0.47 & 0.35 & 0.40 & 0.27 
& 0.15 & 0.02 & 0.22 & 0.11 
& 0.06 & 0.15 & –0.90 \\

VANILLA + Context 
& 0.33 & 0.31 & 0.14 & 0.17 
& 0.13 & 0.01 & 0.21 & 0.07 
& 0.04 & 0.13 & –1.38 \\

FT-1 
& 0.65 & 0.45 & 0.65 & 0.45 
& 0.42 & 0.22 & 0.40 & 0.40 
& 0.17 & 0.39 & 0.46 \\

FT-1 + Context 
& 0.71 & 0.47 & 0.67 & 0.48 
& 0.40 & 0.20 & 0.39 & 0.37 
& 0.16 & 0.37 & 0.40 \\

FT-2 
& \textbf{0.79} & \textbf{0.57} & 0.74 & 0.55 
& \textbf{0.47} & \textbf{0.29} & \textbf{0.47} & \textbf{0.43} 
& \textbf{0.23} & \textbf{0.46} & \textbf{0.51} \\

FT-2 + Context 
& 0.73 & 0.56 & \textbf{0.78} & \textbf{0.58} 
& 0.44 & 0.26 & 0.44 & 0.39 
& 0.19 & 0.40 & 0.37 \\
\hline
\end{tabular}}
\caption{Evaluation of Outcome Prediction and Explanation across Different Fine-tuning Configurations}
\label{tab:outcome_explanation_evaluation}
\end{table*}
To evaluate the effectiveness of our Bail Prediction System, we adopt a comprehensive set of metrics covering both classification accuracy and explanation quality. The evaluation is conducted on two fronts: the bail prediction task and the explanation generation task. We report Precision, Recall, F1, and Accuracy for bail prediction, and we use both quantitative and qualitative methods to evaluate the quality of explanations generated by the model.

\begin{enumerate}
    \item \textbf{Lexical-based Evaluation:} We utilized standard lexical similarity metrics, including Rouge-L~\cite{lin-2004-rouge}, BLEU \cite{papineni-etal-2002-bleu}, and METEOR \cite{banerjee-lavie-2005-meteor} which measure the overlap and order of words between the generated explanations and the reference texts.

    \item \textbf{Semantic Similarity-based Evaluation:} To capture the semantic quality of the generated explanations, we employed BERTScore \cite{BERTScore}, which measures the semantic similarity between the generated text and the reference explanations. 

    \item \textbf{LLM-based Evaluation (LLM-as-a-Judge):} To complement traditional metrics, we incorporate an automatic evaluation strategy that uses large language models themselves as evaluators, commonly referred to as \textit{LLM-as-a-Judge}. This evaluation is crucial for assessing structured argumentation and legal correctness in a format aligned with expert judicial reasoning. We adopt G-Eval~\cite{liu-etal-2023-g}, a GPT-4-based evaluation framework tailored for natural language generation tasks. G-Eval leverages chain-of-thought prompting and structured scoring to assess explanations along three key criteria: \textit{factual accuracy}, \textit{completeness \& coverage}, and \textit{clarity \& coherence}. Each generated legal explanation is scored on a scale from 1 to 10 based on how well it aligns with the expected content and a reference document. The exact prompt format used for evaluation is shown in Appendix Table~\ref{tab: prompt}. For our experiments, we use the GPT-4o-mini model to generate reliable scores without manual intervention. This setup provides an interpretable, unified judgment metric that captures legal soundness, completeness of reasoning, and logical coherence, beyond what traditional similarity-based metrics can offer.


\end{enumerate}



\section{Results and Analysis}

Table~\ref{tab:outcome_explanation_evaluation} presents a comprehensive evaluation of six experimental configurations of the \textit{phi-4} model, varying in terms of fine-tuning and the use of statutory context during inference. These configurations were designed to isolate the contributions of domain-specific supervision and external knowledge augmentation to both prediction accuracy and explanation quality.

The base version of the model, denoted as \texttt{VANILLA}, serves as the starting point. This configuration involves no fine-tuning and no statutory context. Unsurprisingly, its performance is modest across the board, achieving an accuracy of only 0.47. The generated rationales are short, fragmented, and largely generic. This is expected, as the base model lacks prior exposure to the specific structure and semantics of Indian bail judgments and legal reasoning. Interestingly, when statutory text is introduced at inference time in \texttt{VANILLA + Context}, performance drops even further (accuracy declines to 0.33, recall plummets to 0.14, and BLEU and METEOR fall as well). This indicates that the base model not only fails to utilize the additional legal information but is possibly overwhelmed or distracted by the lengthy and complex provisions. It highlights a crucial insight: retrieval-augmented generation (RAG) without task-specific training does not necessarily improve outcomes and may in fact hurt performance in tightly structured legal prediction tasks.

The picture changes significantly with the introduction of fine-tuning. In the \texttt{FT-1} setup, the model is trained on structured features extracted from bail cases (e.g., statutes, facts, arguments, health conditions), allowing it to learn typical decision patterns observed in real judicial outcomes. This leads to a clear jump in all metrics: accuracy improves to 0.65, and explanation quality also rises considerably (BLEU 0.22, METEOR 0.40, BERTScore 0.40). Notably, the model begins to generate meaningful and coherent rationales that are contextually aligned with the facts of the case. This confirms the hypothesis that even without explicit legal grounding, simply training on case features enables the model to internalize a pattern of legal decision-making. When statutory context is added during inference in \texttt{FT-1 + Context}, we observe a further boost, particularly in recall (0.67) and F1-score (0.48). The explanations remain strong, although not significantly better than FT-1. This suggests that once the model has learned how legal decisions are typically made, the additional statutory information serves to fill in edge cases or add confidence, especially in complex or ambiguous scenarios.

The most impressive performance comes from the \texttt{FT-2} configuration, where the model is fine-tuned not only on structured features but also on the meaning and structure of statutes. This exposure allows the model to develop a deeper understanding of how statutory language translates into judicial reasoning. FT-2 achieves the highest overall accuracy (0.79), precision (0.57), and reasoning metrics across all dimensions (BLEU 0.29, METEOR 0.47, BERTScore 0.43). The generated rationales are often detailed, citing relevant sections and interpreting them in the context of case-specific facts, showcasing a level of abstraction and legal alignment beyond mere memorization. These results underscore the value of embedding statutory semantics directly into the model during training, effectively allowing it to learn not just from outcomes but from the legal foundations that justify those outcomes.

Adding RAG-based statutory context at inference time in \texttt{FT-2 + Context} provides the best recall (0.78) and F1-score (0.58), reinforcing the idea that retrieval-augmented inputs help when the model already understands how to interpret them. However, this setup sees a slight drop in precision and explanation quality compared to FT-2, which suggests a possible redundancy or even interference. Once the model already possesses a refined internal representation of legal semantics, external information may sometimes disrupt rather than assist, especially if irrelevant or loosely related statutes are retrieved.

Overall, our results highlight a few key insights. First, fine-tuning is not just helpful; it is essential for legal judgment prediction and explanation. Second, RAG-based statute retrieval is only beneficial when used in conjunction with a fine-tuned model. Third, and most importantly, exposing the model to statutes during training (as in FT-2) yields the most coherent and legally-grounded rationales. This shows the importance of training models not only on case-specific facts but also on the underlying laws they must interpret. In sum, legal AI systems perform best when their internal knowledge is aligned with the legal structure of the domain, and when external context is introduced judiciously and with purpose.

\section{Conclusion and Future Scope}

In this work, we introduced the IBPS, a comprehensive AI framework for predicting bail outcomes and generating fact-based legal rationales. By leveraging a newly curated dataset of over 150,000 High Court bail judgments, we demonstrated the effectiveness of fine-tuning large language models using structured case features and statutory knowledge. Our experiments across various inference configurations highlight that training with legal attributes and statutes leads to more accurate and explainable decisions. The proposed approach not only achieves strong performance on legal prediction tasks but also lays the foundation for transparent and interpretable legal AI systems.

Looking ahead, IBPS can be extended in multiple directions. First, future work may incorporate multi-accused scenarios and additional case complexities such as co-accused interactions and multi-layered charges. Second, the inclusion of judgments in regional Indian languages will enhance linguistic diversity and applicability across different jurisdictions. Lastly, integrating judicial precedent retrieval and assessing real-world deployment through collaborations with courts and legal aid clinics can further validate the practical utility of the system. We believe IBPS marks an important step toward data-driven legal decision support in India.

\section*{Limitations}

While our study presents a large-scale, diverse, and well-structured dataset of Indian High Court bail judgments and leverages advanced fine-tuning techniques over LLMs, several limitations remain that merit consideration. One of the primary challenges lies in the inherent class imbalance present in the dataset. The majority of cases pertain to anticipatory and regular bail applications, with a very small proportion comprising bail cancellation requests. Additionally, the outcome classes are unevenly distributed, with ``Granted'' outcomes significantly outnumbering other classes. This imbalance may skew the model’s learning and evaluation, leading to stronger performance on overrepresented categories while underperforming on underrepresented ones. Though we adopt robust evaluation protocols, future iterations should incorporate data rebalancing or class-aware training techniques to improve generalization across all classes.

Another constraint stems from the scope of training, which is currently restricted to cases involving a single accused. In real-world legal contexts, especially in criminal matters, it is common to encounter cases with multiple co-accused individuals, each with distinct roles, backgrounds, or grounds for bail consideration. Our present model does not handle such scenarios, which limits its applicability in more complex judicial situations. Extending the modeling framework to handle multiple-accused cases is a promising direction for future enhancement.

Finally, the judgments used for training and evaluation were exclusively in English, as they were sourced from High Courts where English is the predominant language of record. However, many bail decisions across Indian states are authored in regional languages. The current language constraint restricts our model’s utility in those jurisdictions and limits its inclusivity. Expanding the dataset and modeling pipeline to support multilingual judgments would be essential to enable broader deployment and fairness across linguistic demographics.

Despite these limitations, our study provides a strong foundational benchmark and modeling framework for the bail prediction and explanation task in the Indian legal domain. We view these constraints not as critical barriers, but as opportunities for future expansion and refinement.

\newpage
\bibliography{anthology, custom, sknigam, legal_bib}

\newpage
\appendix
\clearpage

\appendix
\section{Supplementary Figures and Analyses}

\subsection{Age-Based Outcome Analysis}
\begin{figure}[htbp]
  \centering
  \includegraphics[width=\columnwidth]{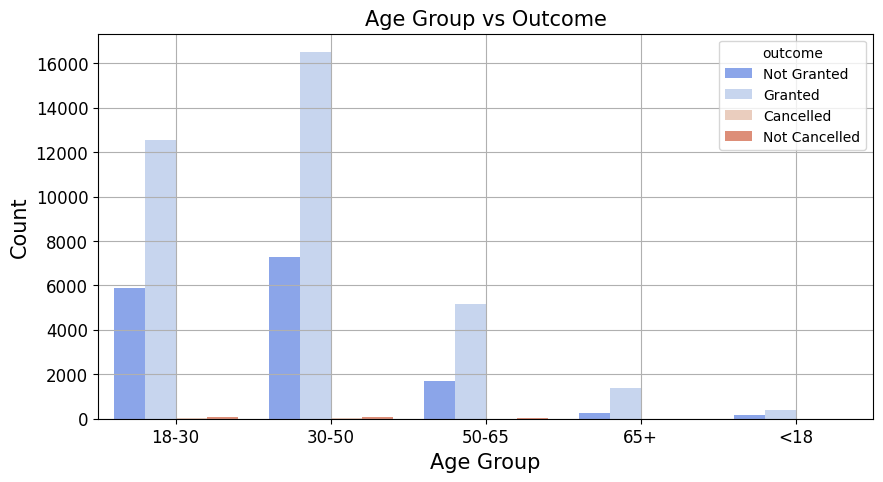}
  \caption{Bail outcomes across different age groups.}
  \label{fig:age_outcome_disparities}
\end{figure}

\noindent
Figure~\ref{fig:age_outcome_disparities} highlights how age influences bail outcomes. Applicants aged 65 and above had the highest grant rate (84.1\%), while those in the 18–30 age group had the lowest (67.7\%). This suggests a potential judicial inclination toward granting bail to older individuals, possibly due to perceived lower flight risk or greater humanitarian concern.

\vspace{1em}
\subsection{Statutes with Lowest Grant Rates}
\begin{figure}[htbp]
  \centering
  \includegraphics[width=\columnwidth]{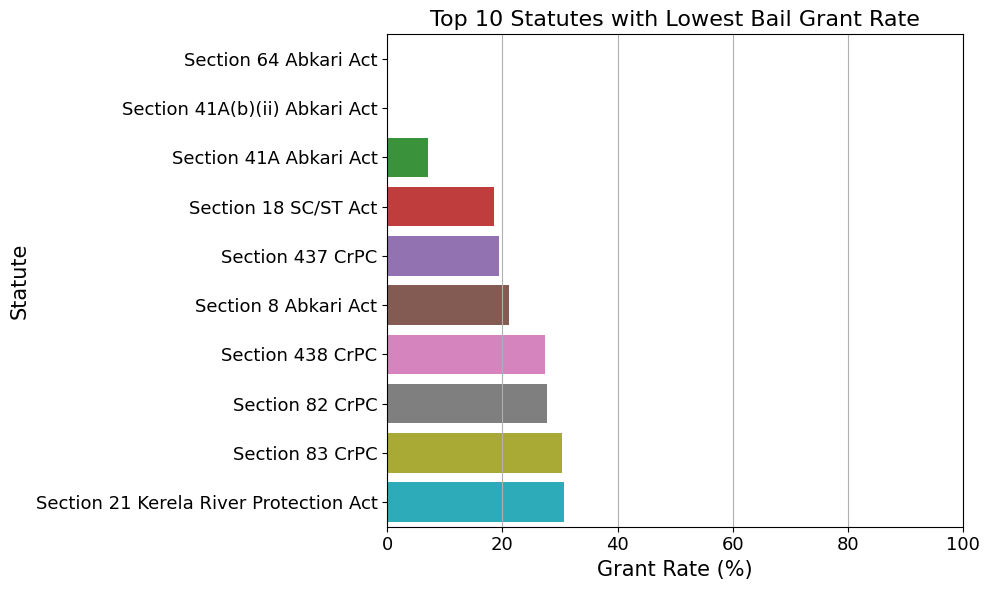}
  \caption{Statutes with lowest bail grant rates.}
  \label{fig:statutes_low_grant_appendix}
\end{figure}

\noindent
Figure~\ref{fig:statutes_low_grant_appendix} presents statutes associated with the lowest bail grant rates. Sections under the Abkari Act (e.g., Section 64, Section 41A(b)(ii)) had a 0\% grant rate. Other socially sensitive laws like Section 18 of the SC/ST Act had grant rates below 20\%, indicating stricter judicial standards for such offenses.

\vspace{1em}
\subsection{Crime-Specific Custody Duration Patterns}
\begin{figure}[htbp]
  \centering
  \includegraphics[width=\linewidth]{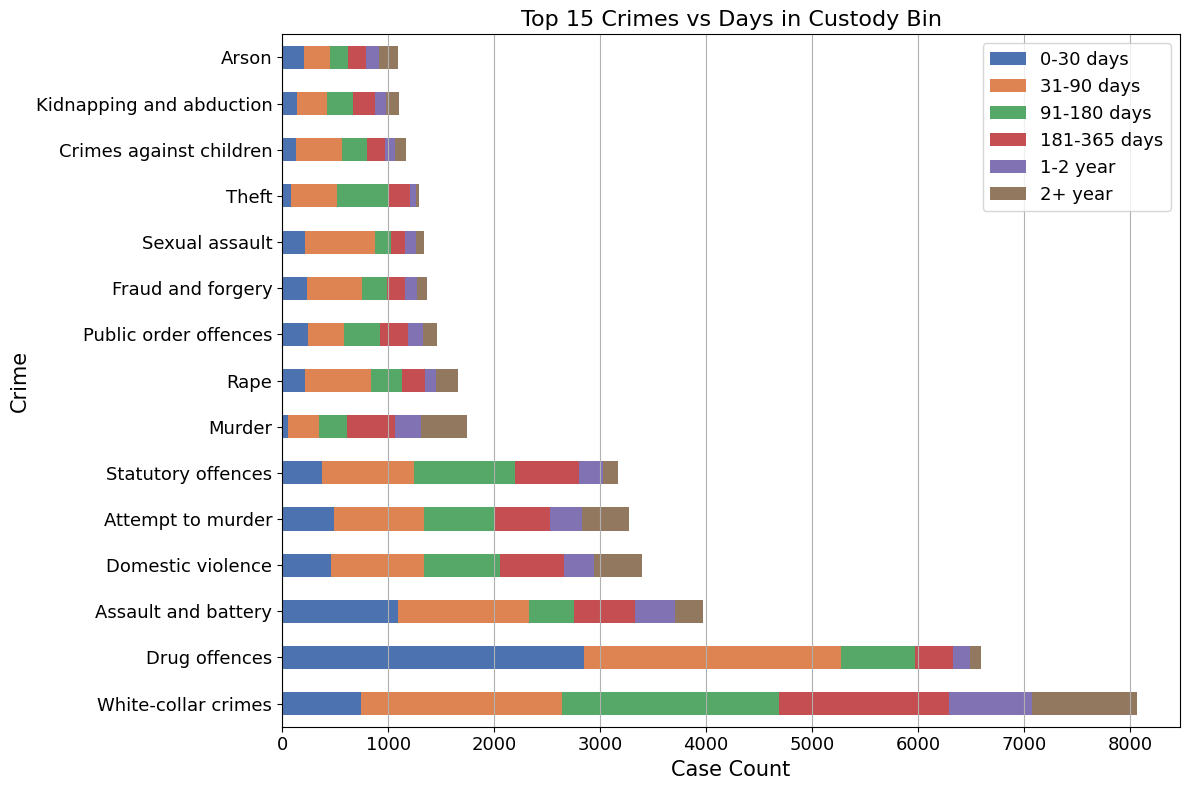}
  \caption{Custody duration distribution by type of crime.}
  \label{fig:crime_custody_appendix}
\end{figure}

\noindent
Figure~\ref{fig:crime_custody_appendix} reveals that murder cases had the longest pretrial detentions, 24.5\% of such cases extended beyond two years. In contrast, drug and sexual assault cases were resolved more quickly, with over 80\% of drug-related cases and 49.9\% of sexual assault cases resolved within 90 days. This highlights disparities in trial speed, possibly linked to the complexity of evidence gathering or legal procedure.

\vspace{1em}
\subsection{Application Withdrawal Rates}

Only 1.3\% of bail applications (2,020 out of 150,430) were withdrawn by the applicants. This low withdrawal rate indicates that the majority of applicants follow through with judicial determination rather than opting out voluntarily. It may reflect confidence in obtaining relief or a lack of viable alternatives.
\begin{figure}[htbp]
  \centering
  \includegraphics[width=\columnwidth]{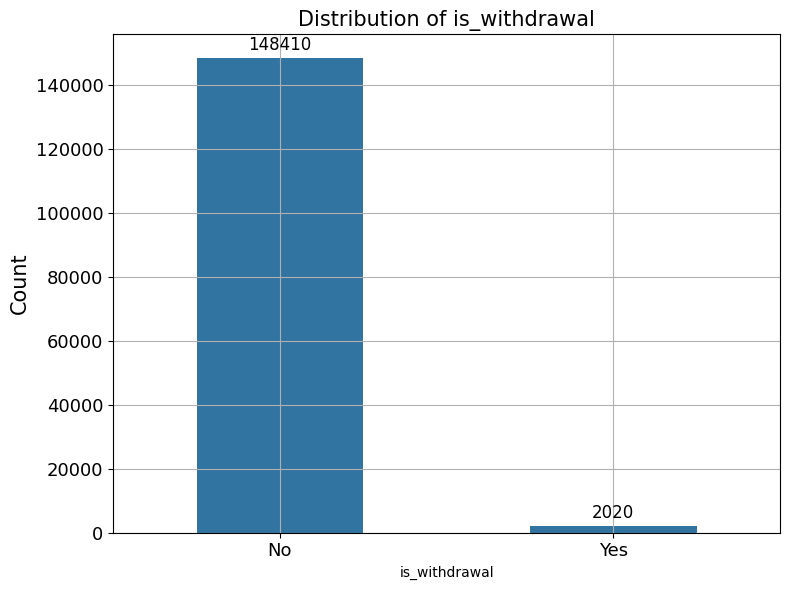}
  \caption{Bar graph showing the no. of withdrwal of bail cases.}
  \label{fig:crime_custody_appendix}
\end{figure}

\vspace{1em}
\subsection{Crime-Age Correlation}
\begin{figure*}[htbp]
  \centering
  \includegraphics[width=\linewidth]{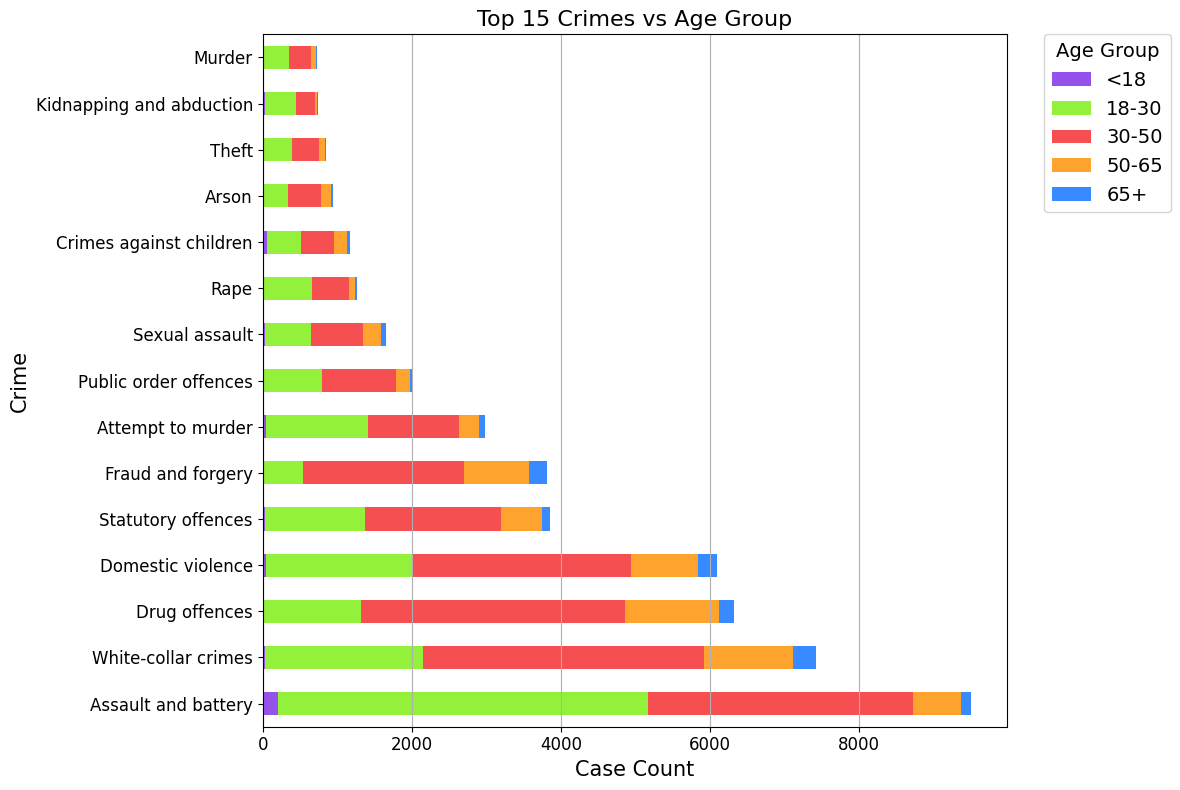}
  \caption{Age distribution of accused across crime categories.}
  \label{fig:crime_age_appendix}
\end{figure*}

\noindent
Figure~\ref{fig:crime_age_appendix} explores age distribution across crime types. Youth (18–30) were predominant in crimes like kidnapping (56.9\%), assault (52.2\%), and rape (51.4\%). Middle-aged individuals (30–50) dominated in drug-related (56.1\%), fraud (56.6\%), and white-collar crimes (50.9\%). These age-crime relationships offer insights into offender profiling and potential rehabilitative strategies.


\begin{figure}[htbp]
  \centering
  \includegraphics[width=\columnwidth]{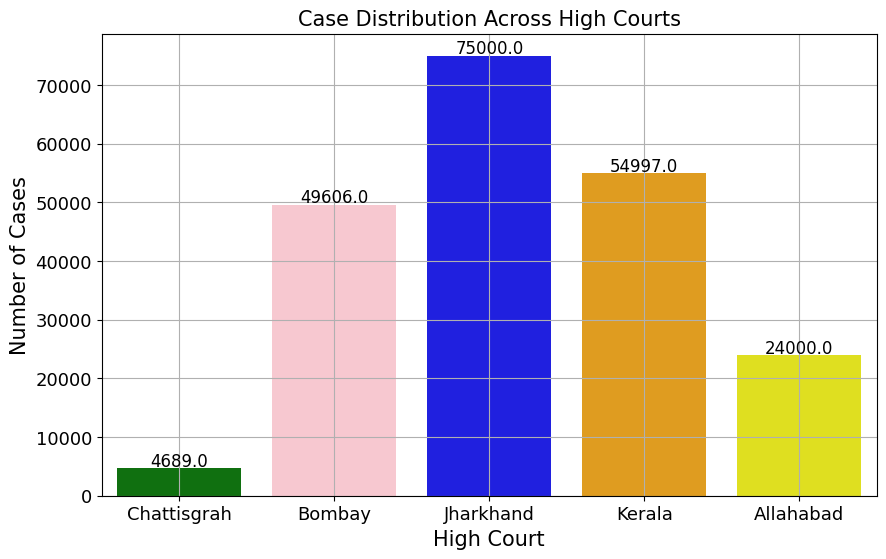}
  \caption{Histogram showing the distribution of cases across various Indian High Courts.}
  \label{fig:case_distribution_courts}
\end{figure}

\begin{figure}[htbp]
  \centering
  \includegraphics[width=\columnwidth]{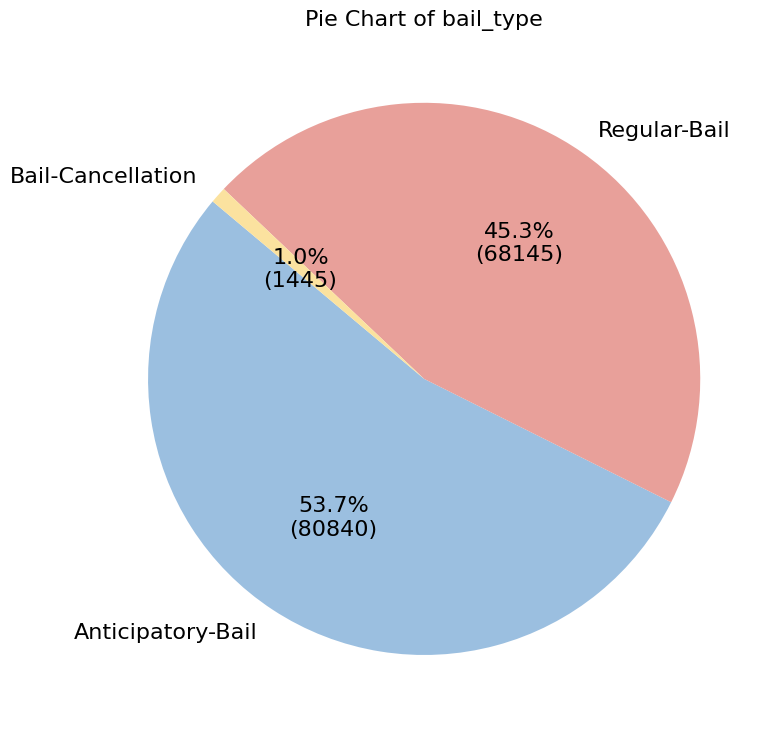}
  \caption{Distribution of bail application types.}
  \label{fig:bail_type_distribution}
\end{figure}

\begin{figure}[htbp]
  \centering
  \includegraphics[width=\columnwidth]{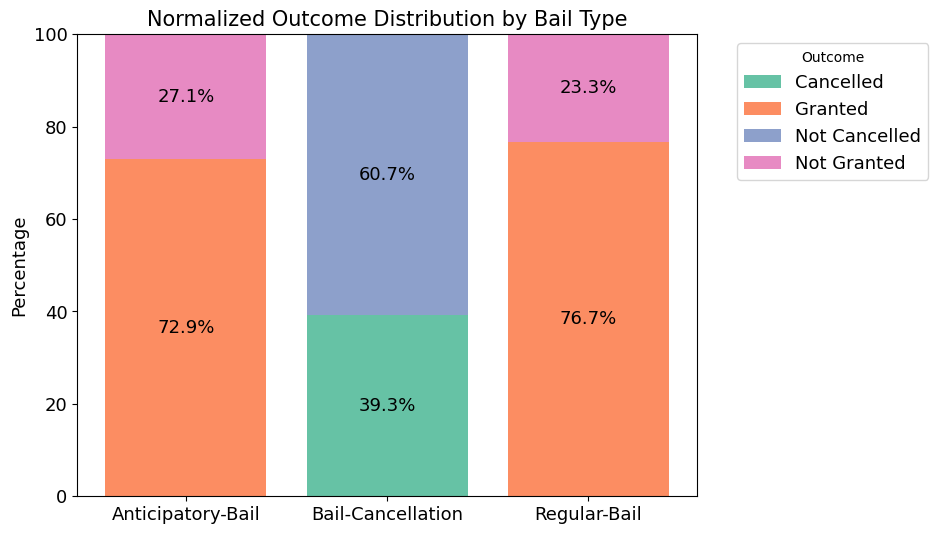}
  \caption{Bail outcome by application type.}
  \label{fig:bail_type_outcome}
\end{figure}

\begin{figure}[htbp]
  \centering
  \includegraphics[width=\columnwidth]{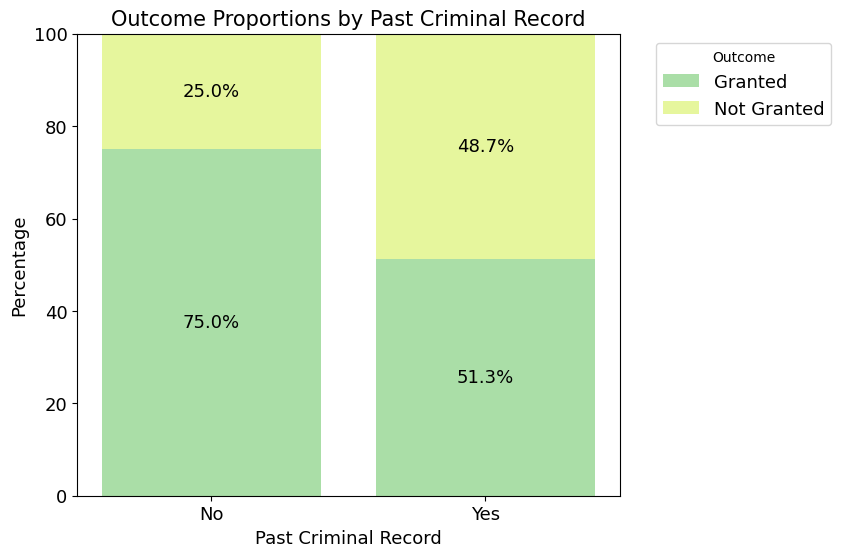}
  \caption{Effect of past criminal record on bail outcome.}
  \label{fig:past_record_outcome}
\end{figure}

\begin{figure}[htbp]
  \centering
  \includegraphics[width=\columnwidth]{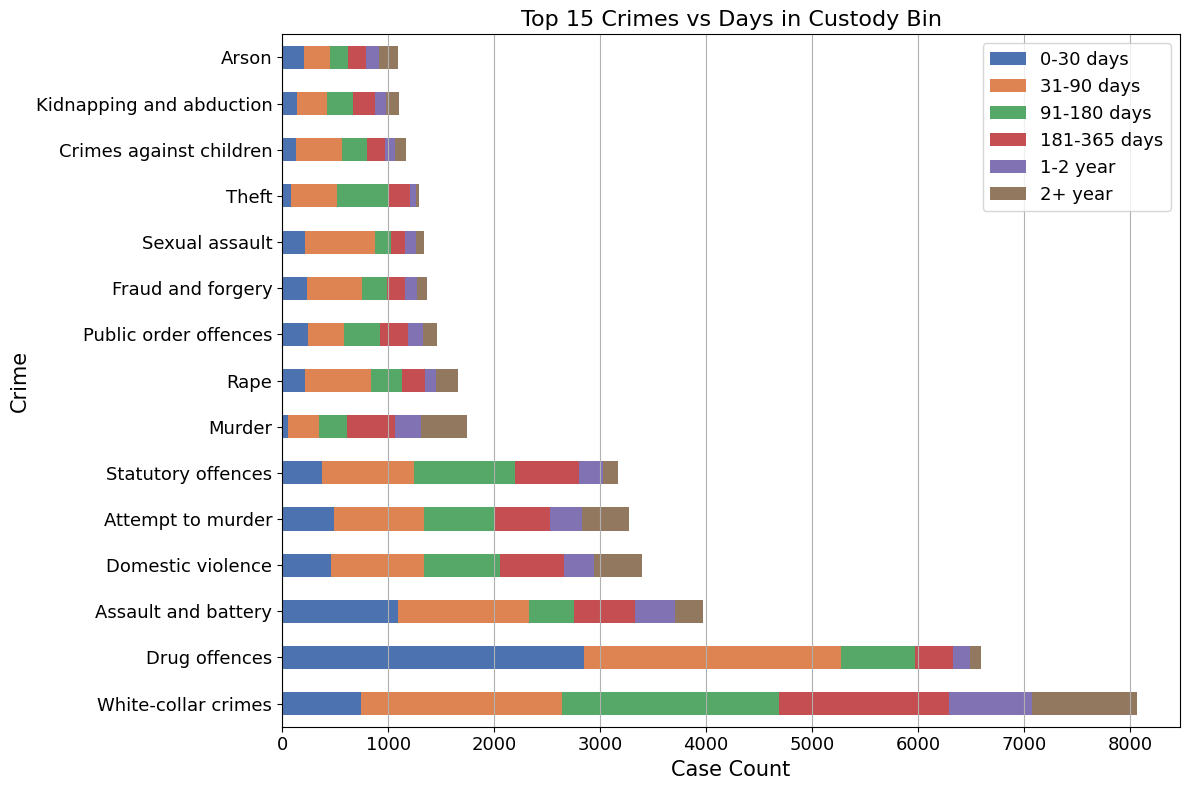}
  \caption{Custody duration by type of crime.}
  \label{fig:crime_custody}
\end{figure}

\begin{table}[!htbp]
  \centering
  \resizebox{\columnwidth}{!}{
      \begin{tabular}{|p{4cm}|p{6cm}|}
        \hline
        \textbf{Model} & \textbf{Limitations} \\ \hline
        en\_legal\_ner\_trf & Weak detection of sections/statutes, lack of mapping between sections and statutes \\ \hline
        BiLSTM-CRF (LegalSeg) & Most sentences were tagged as `None' \\ \hline
        MTL model (LegalSeg) & Slightly better than BiLSTM-CRF, still inadequate \\ \hline
        SparkNLP model (InLegalNER) & Comparable to en\_legal\_ner\_trf, limited by training corpus \\ \hline
        Rule-based Python script & High precision for statute detection, not suitable for other features \\ \hline
      \end{tabular}
  }
  \caption{Feature extraction models and their limitations}
  \label{tab:feature_ext_models}
\end{table}

\begin{figure*}[htbp]
  \centering
  \includegraphics[width=\textwidth]{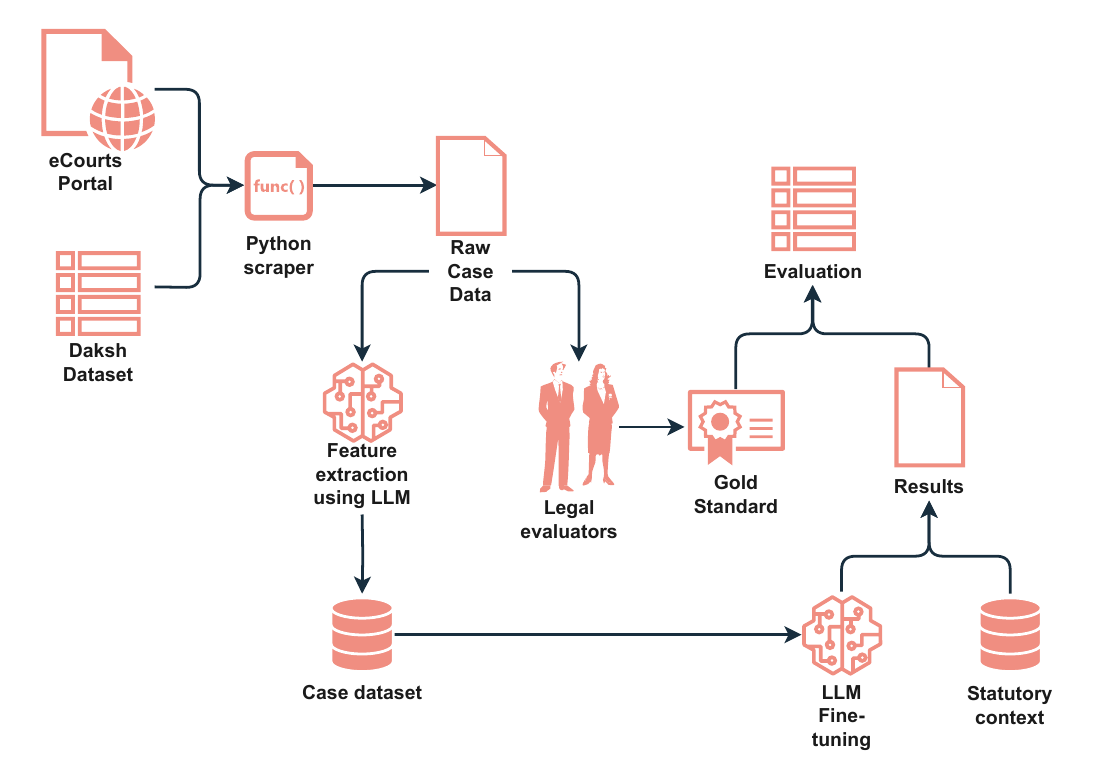}
  \caption{Flowchart representing the IBPS dataset preparation pipeline.}
  \label{fig:fullwidth}
\end{figure*}

\begin{table*}[!ht]  
  \centering
  \small
  \setlength{\tabcolsep}{6pt}
  \begin{tabular}{|l|c|c|c|}
    \hline
    \textbf{Prompt} \\ \hline
    \begin{minipage}{1.0\linewidth}
        \begin{verbatim}   

  
f'''
    given below is a python dictionary format to be filled with information about
    the case given in the raw judgement text. replace the text between the < and > with
    the information extracted from the raw judgement text.
    **DO NOT COPY THE TEXT BETWEEN < AND >, INSTEAD REPLACE IT WITH THE EXTRACTED INFORMATION.**
    
    python dict format :
    {
        "case":"""
            Applicant applied for <type of application, "Regular-Bail" OR "Anticipatory-
            Bail" OR "Bail-Cancellation", (one of these)>.
            Is it a withdrawal application? <"Yes" or "No" depending upon if it is 
            application for withdrawal>.
            Age of the accused is <age of the accused if provided, else write "not
            provided">.
            Health issues for the accused are <description of health issues if provided,
            else "None">.
            There are <"no" if there are no past criminal records, else "some"> past 
            criminal records of the accused.
            Statutes mentioned in the judgement are <list of statutes, 
            eg: [Section 438 CrPC, Section 294(a) IPC, Section 506(1)(b) IPC, Section 34
            IPC, Section 25 Arms Act], 
            do not include the acts/codes/sections that were removed or replaced later>.
            Precedents mentioned in the judgement are <list of precedents, if any, else 
            "None">.
            Details of the incident are <details of the incident if provided, else "None">.
            Arguments supporting the bail application are <arguments supporting the bail
            application, if any, else "None">.
            Arguments opposing the bail application are <arguments opposing the bail 
            application, if any, else "None">."
        """
        "outcome": "The outcome of the case is <status of the outcome, "Bail granted" OR
        "Bail not granted" OR "Bail cancelled" OR "Bail not cancelled", (one of these)>.
        The bail conditions are <list of bail conditions, if any, else "None">.
        "reasoning": "The reasoning for the judgement is <list of reasoning, if any, else "None">."
        "date_of_arrest": "<date of arrest, if provided, else "not provided">."
        "date_of_judgement": "<date of judgement, if provided, else "not provided">."
    }
    
    For example, if the raw judgement text is as follows:

    IN THE HIGH COURT OF KERALA AT ERNAKULAM ...

    the output should be as follows:
    ```json
    {
        "case": "Applicant applied for ...
    }
    ```
    similarly given a raw judgement text, extract the information and convert it into 
    given json format. 
    Respond **ONLY** with valid JSON matching the schema. Do not add explanations or data
    from example itself into the JSON.
    
    Raw Judgement to process: 
    < case to be processed >
    
'''
        \end{verbatim}        
    \end{minipage}
    \\ \hline
  \end{tabular}
\caption{Prompt used across all one-shot/few-shot inference setups for extracting structured case data from raw judgments.}
  \label{tab: prompt}
\end{table*}

\end{document}